\documentclass[10pt,twocolumn,letterpaper]{article}

\usepackage{wacv}
\usepackage{times}
\usepackage{epsfig}
\usepackage{graphicx}
\usepackage{amsmath}
\usepackage{amssymb}

\usepackage{xspace}
\usepackage{multirow}
\usepackage{array}
\usepackage{url}
\usepackage{subcaption}
\usepackage{stmaryrd}
\usepackage{enumitem}
\usepackage{color}
\usepackage{amsmath}
\usepackage{amssymb}
\usepackage{diagbox}
\usepackage[normalem]{ulem}
\usepackage[switch]{lineno}
\usepackage{comment}
\usepackage{multibib}
\usepackage{listings}
\usepackage{authblk}
\usepackage{wacv}
\usepackage{times}
\usepackage{epsfig}
\usepackage{graphicx}
\usepackage{amsmath}
\usepackage{amssymb}
\usepackage{algorithm}
\usepackage{algorithmic}
\usepackage{subcaption}
\usepackage{xcolor}
\usepackage[pagebackref=true,breaklinks=true,letterpaper=true,colorlinks,bookmarks=false]{hyperref}

\newcommand\mdoubleplus{\mathbin{+\mkern-10mu+}}

\newcommand{\OURNAME}{\textit{SAC}\xspace}

\newcommand{\todo}[1]{}

\title{\OURNAME: Semantic Attention Composition for Text-Conditioned Image Retrieval}

\makeatletter
\newcommand{\printfnsymbol}[1]{%
  \textsuperscript{\@fnsymbol{#1}}%
}

    \let\Title\@title
\makeatother

\def\wacvPaperID{1371} 

\wacvfinalcopy 

\ifwacvfinal
\fi

\ifwacvfinal
    \usepackage[breaklinks=true,bookmarks=false]{hyperref}
\else
    \usepackage[pagebackref=true,breaklinks=true,colorlinks,bookmarks=false]{hyperref}
\fi

\newcommand*\samethanks[1][\value{footnote}]{\footnotemark[#1]}

\begin{document}


\author[1]{Surgan Jandial\thanks{equal contribution}\thanks{Work done during the intership at Adobe.}}
\author[2]{Pinkesh Badjatiya\printfnsymbol{1}\thanks{Work done while at Media and Data Science Research Lab, Adobe}}
\author[3]{Pranit Chawla\printfnsymbol{1}}
\author[4]{Ayush Chopra\printfnsymbol{1}\samethanks}
\author[1]{Mausoom Sarkar}
\author[1]{Balaji Krishnamurthy}

\affil[1]{Media and Data Science Research Lab, Adobe}
\affil[2]{Microsoft, India}
\affil[3]{Indian Institute of Technology, Kharagpur}
\affil[4]{Media Lab, Massachusetts Institute of Technology}
\maketitle
\thispagestyle{empty}

\begin{abstract}
The ability to efficiently search for images is essential for improving the user experiences across various products. Incorporating user feedback, via multi-modal inputs, to navigate visual search can help tailor retrieved results to specific user queries. We focus on the task of text-conditioned image retrieval that utilizes support text feedback alongside a reference image to retrieve images that concurrently satisfy constraints imposed by both inputs. The task is challenging since it requires learning composite image-text features by incorporating multiple cross-granular semantic edits from text feedback and then applying the same to visual features.
To address this, we propose a novel framework SAC which resolves the above in two major steps: "where to see" (Semantic Feature Attention) and "how to change" (Semantic Feature Modification). We systematically show how our architecture streamlines the generation of text-aware image features by removing the need for various modules required by other state-of-art techniques.
We present extensive quantitative, qualitative analysis, and ablation studies, to show that our architecture \OURNAME outperforms existing techniques by achieving state-of-the-art performance on 3 benchmark datasets: FashionIQ, Shoes, and Birds-to-Words, while supporting natural language feedback of varying lengths.

\end{abstract}

\section{Introduction}
\label{sec:introduction}
The ability to search for images over an indexed catalog is a fundamental task that serves as a cornerstone for several allied user experiences like smart, intuitive experiences for online commerce such as fine-grained tagging~\cite{finegrained2, attribute}, virtual try-on~\cite{tryon1}, product recommendations~\cite{recommend} and visual search~\cite{similar_search}.
\begin{figure}[ht!]
    \centering
    \includegraphics[page=1,width=0.6\linewidth]{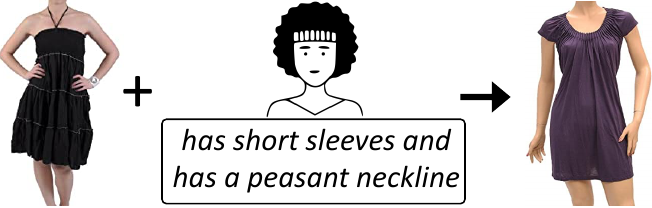}\\
    \includegraphics[page=2,width=0.6\linewidth]{figures/intro_image.pdf}
    \caption{%
        \label{fig:problem-vision}%
        Given a \textit{reference image} and a \textit{support text}, we focus on the task of retrieving images that resemble the \textit{reference image} while also satisfying constraints imposed by the \textit{support text}.%
}%
\end{figure}
The most ubiquitous frameworks in image search either take image or text as input query to search for relevant items \cite{applications_2,imagesearch}. However, a key limitation of these frameworks is the in-feasibility to capture detailed user requirements, either with a single image or a combination of keywords. Correspondingly, several interactive paradigms are being explored, incorporating feedback to help tailor retrieved results to specific user intentions. These interactions involve refining a reference query image through feedback in form of spatial layouts~\cite{spatial}, scene-graphs~\cite{scene_graph, scene2} or relative attributes~\cite{fashion200Kdataset, relative_attribute}. More recently, text feedback via keywords~\cite{TIRG} or short captions~\cite{VALpaper} are being explored to provide more expressive flexibility to the user during interactive image search~\cite{shoes_guo}. This task is denoted as text-conditioned image retrieval. As shown in Figure~\ref{fig:problem-vision}, the task of text-conditioned image retrieval utilizes a support text feedback alongside a reference image with the objective of retrieving image results that can satisfy constraints imposed by both components of the multi-modal input.  \\
Broadly, this task requires learning composite image-text representations that transform only the image features relevant to the text modification while preserving the rest. Several works such as \cite{TIRG}, \cite{VALpaper}, \cite{composeAE} have tried to address this issue. First in this domain, TIRG \cite{TIRG} proposed a simple method leveraging gating and residual modules on the average pooled features from the terminal image layers. Adding to this, ComposeAE \cite{composeAE}, in their work suggests that TIRG assigns huge importance to image features than the text ones, and hence they propose a novel complex space to learn the composition respectively. Although working well to their capacity, the above methods fail to account for wide range of queries and visual concepts and hence are limited in their performance. Moving further, a recent method VAL~\cite{VALpaper} proposed to employ multiple composition modules at varying depths rather only the last. Unlike the previous methods that operate on the average pooled features from the last layer, VAL's composition module transforms the entire Image Volume. However, this not only requires complex series of steps but as VAL perturbs the entire Image Volume, it incurs an additional module just to preserve the features of the input image (as required by the task). Thus, posing challenges to performance and the leveraged compute. Moreover, even though VAL \cite{VALpaper} composes the image and text at varying depths, it does not account for interactions among features across levels of conv layers.

In this work, we propose \OURNAME that resolves the above issues in two major steps: `where to see' and `how to change', and subsequently propose two modules respectively. For example, in Row 1 of Figure \ref{fig:problem-vision}, the support text requires that the modified image has \textbf{short sleeves} and a \textbf{peasant neckline}. This implies that the `where to see' operation should focus on 'sleeves' and `neckline' whereas `how to change' operation should focus on the descriptive attributes of these regions such as `short' and `peasant'. For the first step (`where to see'), we introduce a Semantic Feature Attention (SFA) module which effectively computes the salient regions in the image with respect to the text (i.e. regions which need to be modified). Since CNNs learn visual concepts with increasing abstraction (\cite{visualizing_dnn},\cite{deep}), we thus sample image features over two levels to capture the coarse and fine-grained features. For the second stage, we propose a Semantic Feature Modification (SFM) module that takes as input the two-stage image features along with the text vector to 1) aggregate inter-level features (the coarse and fine-grained features) while ensuring rich representation and 2) modify the resultant according to the text.
Further, to focus on several nuances that arise while training composite (Image-Text) features, we propose a unique composition of loss functions. 

Our contributions can be summarized as follows,

\begin{itemize}
    \item We introduce two modules (SFA and SFM) to break down the task of TCIR into two simple steps 'where to see' and 'how to change'. 
    \item We show how our SFA is able to capture salient image regions mentioned in the query and how our SFM module is able to modify these regions according to the query.  
    \item We perform detailed quantitative and qualitative analysis on 3 benchmark datasets and outperform existing state-of-the-art methods.
\end{itemize}

\section{Related Work}
\label{sec:related_work}
\textbf{Product Search and Image Retrieval} attracts significant research interest due to the diverse practical applicability~\cite{applications_image_retrieval}. Conventional works have utilized uni-modal (image or text) queries to retrieve similar~\cite{GSN} or compatible~\cite{Compatibility_MDSR} images. More recently, we have witnessed a surge in interactive multi-modal techniques that incorporate user feedback to navigate visual search. The user interactions can manifest in form of attributes~\cite{attribute,attribute2}, spatial layouts~\cite{spatial, spatial_TPAMI}, sketches~\cite{sketch1} and text descriptions~\cite{TIRG, VALpaper, shoes_guo}. Owing to the ubiquity in existing search engines and flexibility of articulation, using textual support can facilitate fine-grained specificity in user queries. In this work, we pursue the problem of visual search with textual feedback and propose a framework to efficiently handle unconstrained natural language descriptions of varying lengths.

\textbf{Learning Composite Image-Text Representations} involves jointly processing image and text inputs to capture both these contexts effectively and in a way specific to each task. To review a few, we have 1). Visual Question Answering ~\cite{VQA, patro_question} which uses text semantics to localise the image and further generate the answer, 2). Language Grounding \cite{grounding1, mausoom_ground} which requires spatial localization subject to the input text/phrases , 3). Image-Text matching~\cite{AdvTextMatch, cross_attn_matching} which searches for an Image given the natural language phrase and vice-versa, among many other tasks digesting both the modalities. To this contrast, learning representation in our task involves incorporating text inputs to selectively modify the relevant image features in a way that ensures the preservation of the unaltered features.

\textbf{Text Conditioned Image Retrieval} has been well explored in several recent works \cite{TIRG, VALpaper, composeAE}. First in this domain, TIRG~\cite{TIRG} introduced a residual gating operation to fuse latent image and text embeddings. Citing drawbacks of TIRG, ComposeAE \cite{composeAE} proposed their novel complex space for robust composition of Image and Text Concepts. Moving one step further, a recent state of the art, VAL \cite{VALpaper} proposed to use multiple compositional modules over varying convolutional depths. Briefly, VAL first broadcasts and then fuses the text with image feature to obtain a visiolinguistic representation and then performs self-attention to improve visiolinguistic cues and performs Joint-Attentional Preservation (JAP) to preserve image features. Although effective, VAL in its formulation poses several key limitations: need for complex steps and additional modules affecting performance gains and compute, and in-feasibility to model the interactions/relationship among the features obtained over multiple levels. Addressing the aforementioned, we introduce \OURNAME, which tackles the task of TCIR in two steps, first attending to salient regions in a more structured and simplified way and second, by taking into account both inter-level relationships among features across hierarchy and inter-modal relationship between image and text.
\begin{figure*}[ht!]
    \centering
    \includegraphics[width=0.9\linewidth]{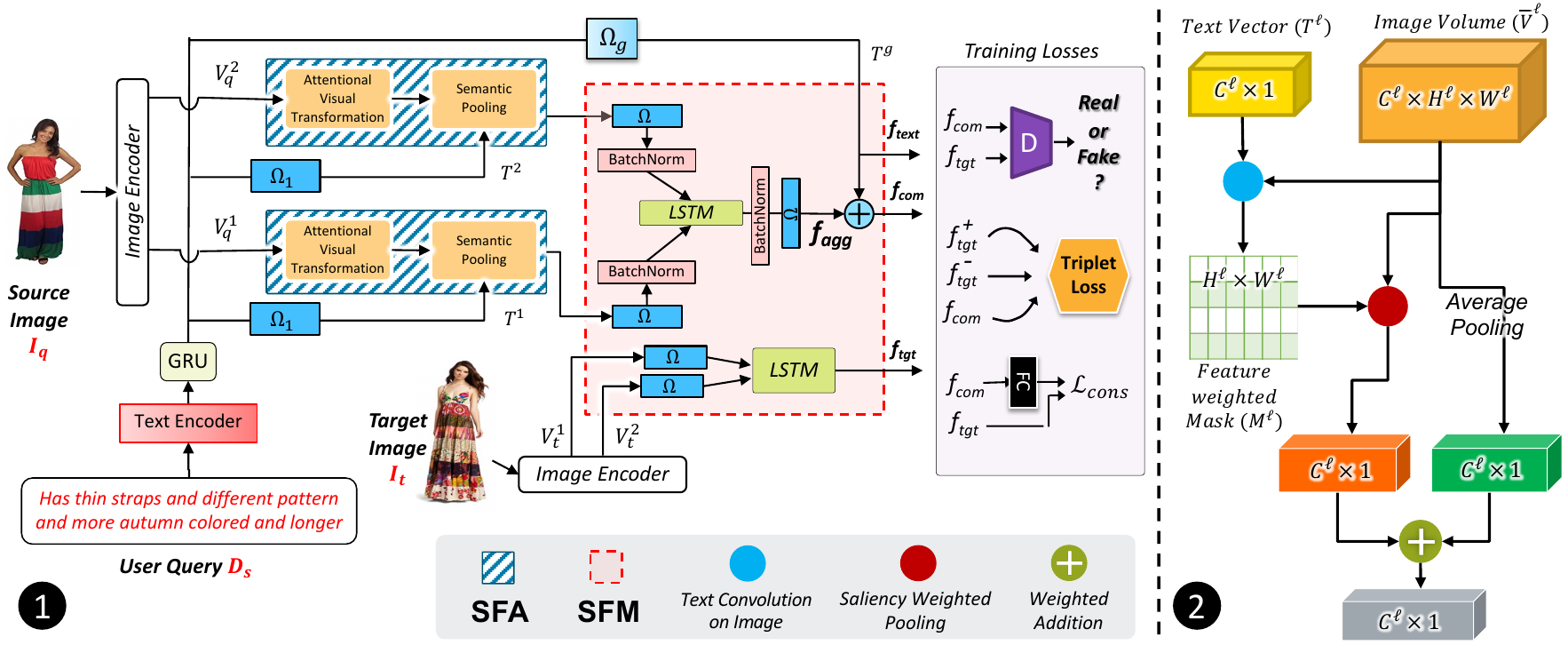}
    \caption{\textbf{(1)} Outline of our proposed \OURNAME framework. We highlight the 2 main components (shaded): the \textbf{Semantic Feature Attention} module, \and the \textbf{Semantic Feature Modification (SFM)} module. \textbf{(2)} Schematic representation of the Semantic Pooling component. Some of the operations are denoted by symbols with their description provided in the legend.}%
    \label{fig:main-architecture}%
    \label{fig:tcfs}%
\end{figure*}
\section{Approach}
\label{sec:approach}
Given a query image ($I_{q}$) and the modification text description ($D_{s}$), the training objective of our task is to learn a Image-Text composite representation that uniquely aligns with the visual representation of the target ground-truth image ($I_{t}$). To debrief our overall approach and the components underlying, we divide this section as:
Section~\ref{subsec:twosteps} presents an overview and motivation behind our approach, which is then followed by each part of our methodology. Section~\ref{sec:approach:encode} provides our strategy to independently encode the image ($I_{q}$, $I_{t}$) and text ($D_{s}$) inputs. Further, Sections~\ref{sec:approach:sfa} to \ref{sec:approach:hfm} delineates different phases of \textbf{\OURNAME}, thus generating visual representation for $I_{t}$ and the composite Image-Text representation for $I_{q}$ and $D_{s}$. In the last part of this section, we present our unique composition of loss functions (in Sec.~\ref{sec:approach:losses}) which are designed to regularize the visual and linguistic features in the composite representations. An overview of the proposed approach is provided in Fig.~\ref{fig:main-architecture}.

\subsection{Learning in Two Stages}
\label{subsec:twosteps}
We intuitively break down the learning of \OURNAME in two stages: \textit{`where to look'} and \textit{`how to change'}.

In the first stage, we utilize the \textbf{Semantic Feature Attention (SFA)} module to find the salient image regions with respect to the text. On the other hand, VAL~\cite{VALpaper} in their very first step fuses the image and text features to obtain visiolinguistic representations and further introduces two set of modules: self-attention to improve the obtained visiolinguistic cues and ``Joint-Attentional Preservation'' (JAP) to preserve the image features which do not have to be modified.
Intuitively, they first learn the coarse level visiolinguistic Image Text Relationships which further undergo transformation and preservation to obtain final features required for retrieval. 

In contrast to learning any complex visiolinguistic transformation, the only task for our Semantic Feature Attention (SFA) is \textit{given an Image, generates the 2-d probability map that describes the importance of each pixel with respect to text}. Therefore, we use the text vector as a kernel and convolve the entire image volume to obtain a probability matrix. The probability matrix is then applied back to the image volume to reweigh the features in accordance with the text importance, hence \textit{`where to look'}. Intuitively, our method keeps the image volume intact and only alters the regions of interest from the original image volume, thus eliminating the need for a separate preservation module (as used in VAL).

\textbf{To handle the `how to change'}, we propose a Semantic Feature Modification (SFM) Module. Since, we sample two levels of image features, the input to SFM includes two re-weighted image features and the text feature vector, which are encapsulated in our novel way to capture relationships across feature levels (coarse and fine) and across modalities (image and text).

\subsection{Representing Image and Text}
\label{sec:approach:encode}
\textbf{Image Encoder:} 
CNNs are well known to encode visual concepts with increasing abstraction, generally, becoming finer as we progress over levels. Following the similar idea, in our method, we propose to sample out two granularities of embeddings: Low-Level features and High-Level features. Furthermore, in Section~\ref{sec:ablations} we provide an analysis on levels as used in our method and related efforts.
Concretely, the resultant visual features $\mathcal{F}_{q}$ and $\mathcal{F}_{t}$ for the query ($I_{q}$) and the target image ($I_{t}$) respectively, are computed as,%
\begin{equation}
    \begin{split}
        \mathcal{F}_{q} = {\{V_{q}^1, V_{q}^2}\} = \phi_\text{CNN}(I_{q}) \\
        \mathcal{F}_{t} = {\{V_{t}^1, V_{t}^2}\} = \phi_\text{CNN}(I_{t}) \\
    \end{split}
\end{equation}

\textbf{Text Encoder:} To generate text embeddings corresponding to the visual features at the two granularities we use a GRU~\cite{GRU} followed by \textit{2} parallel fully connected layers.
Given the support text $D_{s}$ (max $N$ words), we obtain a sequence of word-level embedding features $F_{word} \in \mathbb{R}^{1 \times 768}$ which are then passed through a GRU to obtain the support text feature $\mathcal{F}_{sent} \in \mathbb{R}^{1 \times 1024}$ as
\begin{equation}
    \mathcal{F}_{sent} = \text{GRU}([F_{word}^{1},~F_{word}^{2},~\cdots,~ F_{word}^{N}])
\end{equation}
We then transform the $\mathcal{F}_{sent}$ through two separate linear projection layers as,%
\begin{equation}
    \mathcal{T}^{1},~\mathcal{T}^{2}  = \Omega_{1}(\mathcal{F}_{sent}),~~   \Omega_{2}(\mathcal{F}_{sent})
\end{equation}

\subsection{Semantic Feature Attention (SFA)}
\label{sec:approach:sfa}
As previously mentioned, the goal of our SFA module is to highlight salient regions in the image which need to be modified according to the text. 
Our SFA Module is made of two major sub-parts: (1) Attentional Visual Transformation (2) Semantic Pooling. Intuitively, the first one captures the importance of a pixel, subject to the other positional locations within the Image, while the second one captures the importance with respect to text. Formally, we define both these operations below. Since SFA at both levels follows the same operations, we use level $\ell$ in the discussion below for brevity.

\textbf{\underline{Attentional Visual Transformation}:} 
Here, we capture apriori long-range contextual relationships within the visual embedding ($V_{q}^{\ell}$) to help enhancing the representational capabilities. Therefore, we leverage a positional attention mechanism to aggregate the spatial context, \cite{sagan,danet} to transform $V_{q}^{\ell}$ into volumetric representation $\overline{V}_{q}^{\ell} \in \mathbb{R}^{C_{\ell} \times H_{\ell} \times W_{\ell}}$. For this, $V_{q}^{\ell}$ is passed through parallel convolutional layers (denoted by $\Theta_q$, $\Theta_k$, $\Theta_v$) and the obtained volume is reshaped to obtain new query and key feature maps denoted by $(Q^{\ell}, K^{\ell}) \in \mathbb{R}^{C_{\ell} \times N_{\ell}}$, $N_{\ell} = H_{\ell} \times W_{\ell}$ which are then used to obtain a spatial attention map $\mathbf{\mathcal{A}}^{\ell}_{\mathbf{self}}$.
\begin{gather*}
    \mathcal{Q}^{\ell} = \Theta_{q}(V_{q}^{\ell}),\quad \mathcal{K}^{\ell} = \Theta_{k}(V_{q}^{\ell}),\quad \mathcal{V}^{\ell} = \Theta_{v} (V_{q}^{\ell}) \\
    \mathbf{\mathcal{A}}^{\ell}_{\mathbf{self}} = \text{softmax}(({\mathcal{Q}^{\ell}})^{T} {\mathcal{K}^{\ell}})
\end{gather*}\vspace{-2pt}%
We generate an intermediate feature $E^{\ell}$ to compute the transformed attentive visual feature map $\overline{V}_{q}^{\ell}$ as,
\begin{equation}
    E^{\ell} = \mathcal{V}^{\ell} ({\mathbf{\mathcal{A}}^{\ell}_{\mathbf{self}}})^T \quad\text{and}\quad \overline{V}_{q}^{\ell} = \beta E^{\ell}  + V_{q}^{\ell}
\end{equation}
The feature vector $\overline{V}_{q}^{\ell}$ encodes global visual information along with selectively aggregated spatial context which improves the semantic consistency in the representation. 

\underline{\textbf{Semantic Pooling:}}
\label{sec:approach:vision-linguistic-pooling}
Further, the learnt attentive visual representation $\overline{V}_{q}^{\ell}$ is now convolved with the corresponding text representation $T^{\ell}$ to obtain a \textit{2-D saliency map)}  $\mathbf{\mathcal{A}}^{\ell}_{\mathbf{sal}} \in \mathbb{R}^{H_{i} \times W_{i}}$. that essentially gives the importance of each pixel with respect to Text.%
\begin{equation}%
    \mathbf{\mathcal{A}}^{\ell}_{\mathbf{sal}} = \overline V_{q}^{\ell} \circledast T^{\ell}
\end{equation}
$\mathcal{A}^{\ell}_{sal}$ is then passed through softmax, with temperature $\mathcal{T}$, to obtain feature-weightage map (probability map) $M^{\ell}$.
\begin{equation}
    M^{\ell} = \text{softmax}({\mathbf{\mathcal{A}}^{\ell}_{\mathbf{sal}}}/{\mathcal{T}})
\end{equation}
We provide a clear representation of the operations performed in Figure~\ref{fig:tcfs}. \\

We then use the obtained feature-weighted map $M^{\ell}$ to pool each channel in the attentional visual feature map $\overline{V}_{q}^{\ell}$ to select image features salient to text features, thus generating $S^{\ell} \in \mathbb{R}^{C_{\ell} \times 1}$ given as,%
\begin{equation}%
    \begin{split}
        S^{\ell}(c) = \sum_{h=1}^{H_{\ell}}\sum_{w=1}^{W_{\ell}} M^{\ell}(h,w)~\circledast~& \overline{V}_{q}^{\ell}(c, h, w)
    \end{split}
\end{equation}
where $1 \leq c \leq \mathcal{C}^{\ell}$ and $\mathcal{C}^{\ell}$ denotes the number of channels.

Finally, the \textit{granular} text-conditioned visual embedding $O^{\ell}_{q}$ is obtained by a weighted addition of the Text Conditioned Image feature $S^{\ell}$ with pooled attentive visual feature map (we use generalized-mean pooling technique \textbf{GeM}~\cite{gem}). The pooled visual embedding for the target image is also obtained as,%
\begin{equation}%
    O^{\ell}_{q} = \textit{Pool}(\overline{V}_{q}^{\ell}) + \gamma S^{\ell} \quad\text{and}\quad O^{\ell}_{t} = \textit{Pool}(V^{\ell}_{t})
\end{equation}
The obtained embeddings for the query image $O^{\ell}_{q}$ and the target image $O^{\ell}_{t}$, across the \textit{two} levels combined, form the resultant salient feature set $\mathcal{F}_{q}^{img}$ and $\mathcal{F}_{t}^{img}$ which is passed on to SFM.
\begin{equation}
    \mathcal{F}_{q}^{img} = \{ O^{1}_{q},O^{2}_{q}\}\quad\text{and}\quad \mathcal{F}_{t}^{img} =  \{ O^{1}_{t},O^{2}_{t}\}
\end{equation}

\subsection{Semantic Feature Modification (SFM)}
\label{sec:approach:hfm}
Here, we address the task of ``how to change'' by compositing the transformed image features with text features.
Inputs to this are salient feature set $\mathcal{F}_{q}^{img}$ and text feature $\mathcal{F}_{sent}$.
Briefly, in this step, we first perform a gating operation over feature levels and then subsequently modify the resultant with text. Formally, we take the feature set $\mathcal{F}_{q}^{img}$ and pass it through two independent linear projections. As features across levels encode different properties and consequently exhibit different output sizes, projecting them to a common space before modeling their interactions is helpful,
\begin{align}
    G_{q}^{1}, G_{q}^{2} = \Omega_{1}([\mathcal{F}_{q}^{img}]_{1}), \Omega_{2}([\mathcal{F}_{q}^{img}]_{2})%
    \label{eq:linrnn}\\
    G_{q}^{1}, G_{q}^{2} = BatchNorm(G_{q}^{1}), BatchNorm(G_{q}^{2})%
    \label{eq:batchrnn}
\end{align}
To selectively pass on features from the lower-level ($G_{q}^{1}$) to higher-level of hierarchy ($G_{q}^{2}$), we further use our gating operation, that uses an LSTM followed by a $BatchNorm$ to obtain aggregated feature vector $f_{agg}$: 
\begin{equation}
    \begin{split}
        H_{} = {LSTM}_{}([G^{1}_{q},{G}^{2}_{q}]) \\
        f_{agg} = \Omega_{}(BatchNorm(H_{})) \\
    \end{split}%
    \label{eq:gated-aggregation}
\end{equation}
To obtain embedding $f_{tgt}$ for the target feature set $\mathcal{F}_{t}^{img}$, we follow the same pipeline of projection up until gating across levels. ($G_{t}^{1},~G_{t}^{2}$ below are obtained using Eq.~\ref{eq:linrnn} and \ref{eq:batchrnn} for $\mathcal{F}_{t}^{img}$)
\begin{equation}
        f_{tgt} = {LSTM}_{}([G_{t}^{1},G_{t}^{2}])
\end{equation}
Next, we take the aggregated feature vector ($f_{agg}$) and the global text vector $f_{text}$ obtained by taking the linear projection $f_{text}=\Omega_{g}(\mathcal{F}_{sent})$. The final composed image-text representation which includes the modifications is then obtained by Residual Offsetting of $f_{agg}$ with $f_{text}$ followed by vector normalization as:
\begin{equation}
    f_{com} = \delta \frac{f_{agg} + f_{text}}{{\lVert{f_{agg} + f_{text}}\rVert}_{2}}
    \label{eq:residual-composition}
\end{equation}
where $\delta$ parameter denotes the learnable normalization scale and ${\lVert .\rVert }_{2}$ denotes the $L_{2}$ norm. We discuss the impact of this residual composition strategy in Section~\ref{sec:ablations}.

$f_{com}$, $f_{tgt}$ and $f_{text}$ are used as inputs to the loss functions detailed in the next section.

\begin{table*}[tbp!]
    \centering
    \scriptsize
    \begin{tabular}{|c||c|c|c|c|c|c|c|c|c|}
        \hline
        \multirow{3}{*}{\diagbox{Method}{Dataset}} & \multicolumn{9}{c|}{FashionIQ} \\
        \cline{2-10}
         & \multicolumn{2}{c|}{Dress} & \multicolumn{2}{c|}{Toptee} & \multicolumn{2}{c|}{Shirt} & \multicolumn{2}{c|}{\textbf{Average}} & \multirow{2}{*}{\textbf{Average}}\\
        \cline{2-9}
        & R@10 & R@50 & R@10 & R@50 & R@10 & R@50 & R@10 & R@50 &  \\ 
        \hline

        Image Only & 2.92 & 10.10 & 4.53 & 11.63 & 5.34 & 14.62 & 4.26 & 12.12 & 8.19 \\ \hline
        Text Only & 8.67 & 25.08 & 9.68 & 28.25 & 8.30 & 25.02 & 8.88 & 26.11 & 17.50 \\ \hline    
        Concat & 9.06 & 27.27 & 10.45 & 29.83 & 9.66 & 28.06 & 9.72 & 28.39 & 19.56 \\ \hline\hline

        FiLM~\cite{film} & 14.23 & 33.34 & 17.30 & 37.68 & 15.04 & 34.09 & 15.52 & 35.04 & 25.28 \\ \hline
        TIRG~\cite{TIRG} & 14.87 & 34.66 & 19.08 & 39.62 & 18.26 & 37.89 & 17.40 & 37.39 & 27.40\\ \hline
        Relationship~\cite{Relationship}& 15.44 & 38.08 & 21.10 & 44.77 & 18.33 & 38.63 & 18.29 & 40.49 & 29.39 \\ \hline
        VAL~\cite{VALpaper} & 21.47 & 43.83 & 26.71 & 51.81 & 21.03 & 42.75 & 23.07 & 46.13 & 34.60 \\ \hline
        VAL w/ GloVe~\cite{VALpaper} & 22.53 & 44.00 &  27.53 & 51.68 & 22.38 & 44.15 & 24.14 & 46.61 & 35.38 \\ 
        \hline\hline
        CurlingNet~\cite{yu2020curlingnet} (FashionIQ-W 2019) & 24.44 & 47.69 & 25.19 & 49.66 & 18.59 & 40.57 & 22.74 & 45.97 & 34.36 \\ \hline
        RTIC~\cite{shin2020fashion} (FashionIQ-W 2020) & \textbf{28.21} & 51.41 & 28.00 & 55.58 & 21.30 & 44.80 & 25.83 & 50.59 & 38.22 \\ \hline
        ComposeAE w/ Random Emb.~\cite{composeAE} (WACV 2021) & 11.99& 31.38 & 11.01	 & 27.48 		& 11.04 & 26.49  & 11.34 & 28.45 & 19.89 \\ \hline
        ComposeAE w/ BERT.~\cite{composeAE} (WACV 2021) &  14.03 & 35.1 & 15.8 & 39.26 & 13.88	& 34.59	 & 19.89 & 36.31 & 25.44 \\ \hline\hline
        
        \textbf{\OURNAME} w/ \textit{Random Emb.} & 26.13 & \textbf{52.10} & 31.16 & 59.05 & 26.20 & 50.93 & 27.83 & 54.03 & 40.93 \\ 
        \textbf{\OURNAME} w/ BERT & 26.52 & 51.01 & \textbf{32.70} & \textbf{61.23} & \textbf{28.02} & \textbf{51.86} & \textbf{29.08} & \textbf{54.70} & \textbf{41.89} \\ \hline
    \end{tabular}
    \caption{Quantitative comparison on FashionIQ dataset. \OURNAME outperforms existing methods using both randomly and BERT-pretrained initialized text embedding. Best numbers are highlighted in \textbf{bold}.}
    \label{tab:results:fashionIQ}\vspace{-10pt}
\end{table*}
\begin{table}[tbp!]
    \centering
    \scriptsize
    \begin{tabular}{|c||c|c|c|}
        \hline
        \multirow{2}{*}{\diagbox{Method}{Dataset}} & \multicolumn{3}{c|}{Birds-to-Words} \\
        \cline{2-4}
        & R@10 & R@50 & \textbf{Average}\\ 
        \hline
        Text Only & 1.69 & 8.34 & 5.01 \\ \hline    
        Image Only & 15.45 & 32.14 & 23.80 \\ \hline
        Concat & 12.05 & 34.27 & 23.16 \\ \hline\hline
        TIRG~\cite{TIRG} & 15.8 & 38.65 & 27.22\\ \hline
        VAL~\cite{VALpaper} & - & - & -\\ \hline
        ComposeAE w/ Random Emb. ~\cite{composeAE} & 10.94 & 29.35 & 20.14\\ \hline
        ComposeAE w/ BERT. ~\cite{composeAE} & 10.66 & 34.84 & 22.75 \\ \hline\hline
        \textbf{\OURNAME} w/ \textit{Random Emb.} & \textbf{20.34} & 44.94 & \textbf{32.64}\\ 
        \textbf{\OURNAME} w/ BERT & 19.56 & \textbf{45.24} & 32.40\\ \hline
    \end{tabular}
    \caption{Quantitative comparison on Birds-to-Words dataset. \OURNAME outperform existing methods using both randomly and BERT-pretrained initialized text embedding.}
    \label{tab:results:birds2words}
\end{table}
\begin{table}[tbp!]
    \centering
    \scriptsize
    \begin{tabular}{|c||c|c|c|c|}
        \hline
        \multirow{2}{*}{\diagbox{Method}{Dataset}} & \multicolumn{4}{c|}{Shoes} \\
        \cline{2-5}
        & R@1 & R@10 & R@50 & \textbf{Average}\\ 
        \hline
        Text Only & 0.60 & 6.20 & 19.42 & 8.74\\ \hline 
        Image Only & 6.07 & 25.6 &  47.87 & 26.51 \\ \hline
        Concat & 5.70 & 20.32 & 39.97 & 22.00 \\ \hline\hline
        FiLM~\cite{film} & 10.19 & 38.39 & 68.30 & 38.96\\ \hline
        TIRG~\cite{TIRG} & 12.60 & 45.45 & 69.39 & 42.48\\ \hline
        Relationship~\cite{Relationship} & 12.31 & 45.10 & 71.45 & 42.95\\ \hline
        VAL (2 level) ~\cite{VALpaper}& 14.98 & 47.25 & - & -\\ \hline
        VAL~\cite{VALpaper} & 16.98 & 49.83 & 73.91 & 46.91\\ \hline
        VAL w/ GloVe~\cite{VALpaper} & 17.18 & 51.52 & 75.83 & 48.18\\ \hline
        ComposeAE w/ Random Emb.~\cite{composeAE} & 3.46 & 20.84 & 52.58 & 25.62 \\ \hline
        ComposeAE w/ BERT ~\cite{composeAE} & 4.37 & 19.36 & 47.58 & 23.77 \\ \hline\hline
        \textbf{\OURNAME} w/ \textit{Random Emb.} & 18.11 & \textbf{52.41} & 75.42 & 48.64 \\ 
        \textbf{\OURNAME} w/ BERT & \textbf{18.5} & 51.73 & \textbf{77.28} & \textbf{49.17} \\ \hline
    \end{tabular}
    \caption{Quantitative comparison on Shoes datasets. \OURNAME outperform existing methods using both randomly and BERT-pretrained initialized text embedding.}
    \label{tab:results:shoes}
\end{table}

\subsection{Loss Functions}
\label{sec:approach:losses}
The training dataset ($I_{train}$) is characterised by $3$-tuples consisting of ($I_{q}, D_{s}, I_{t}$). Correspondingly, $f_{com}$ represents the composed text-conditioned image embedding for ($I_{q}, D_{s}$), $f_{text}$ represents the latent embedding for $D_{s}$ and $f_{tgt}^{+}$ represents the latent embedding for $I_{t}$. Consider another image $I_{n}$ sampled from $I_{train}$, s.t. $I_{n} \notin \{ I_{q} \cup I_{t}\}$ where $f^{-}_{tgt}$ represents its latent visual embedding which is generated using the same pipeline as for $f_{tgt}$. We next explain the different loss functions used to train \OURNAME.

\textbf{Triplet Loss} is the primary training objective which seeks to constrain the \textit{anchor} $f_{com}$ to align with the \textit{target} $f_{tgt}^{+}$ by simultaneously contrasting with the embedding for a \textit{negative} image $f_{tgt}^{-}$. The loss function is defined as%
\begin{equation}%
    \mathcal{L}_{triplet} = \log(1+e^{{\lVert  f_{com} - f_{tgt}^{+} \rVert }_{2}~-~{\lVert  f_{com} - f_{tgt}^{-}\rVert }_{2}})
\end{equation}
where ${\lVert .\rVert }_{2}$ operator denotes the $L_{2}$ norm.

To help learn discriminative representations, we employ a hard negative strategy that interleaves the random selection of $I_{n}$ with an online distance-based sampling technique. This sampling weighs each $I_{n} \in I_{train}$ using the \textit{$L_{2}$}-distance of the corresponding embedding ($f_{tgt}^{-}$) with $f_{com}$ with smaller distances weighted higher. 

\textbf{Discriminator Loss} helps improve the alignment of $f_{com}$ with $f_{tgt}$ by utilizing a discriminator that penalizes distributional divergence of linear projections of these embeddings.%
\begin{equation}%
    \mathcal{L}_{disc} = - \mathbb{E} \big[log(\mathcal{D}(f_{tgt})\big] -  \mathbb{E}\big[log(1-\mathcal{D}(f_{com}))\big]
\end{equation}
where $\mathcal{D}$ is the discriminator network which has three fully-connected layers and is trained end-to-end along with the entire model. Details about the architecture of the discriminator is provided in Appendix~\ref{appendix:implementation_details}.
We also discuss the particular impact of using this discriminator loss in Section~\ref{sec:ablations:loss_functions:discriminator_loss}.

\textbf{Consistency Loss} constraints visual and linguistic projections of $f_{com}$, denoted by $f_{gen}^{img}$ and $f_{gen}^{text}$, to align with latent embeddings $f_{tgt}$ and $f_{text}$ respectively. This objective by reconstruction regularizes and reinforces the balanced utilization of both text and image in composed embedding $f_{com}$.%
\begin{equation}%
    \mathcal{L}_{cons} = \alpha_{t}{\lVert  f_{gen}^{text} - f_{text}\rVert }_{2} ~+~ \alpha_{i}{\lVert  f_{gen}^{img} - f_{tgt} \rVert }_{2}
\end{equation}
where, ${\lVert .\rVert }_{2}$ is the $L_{2}$ norm.

In the above equation, we project the vector $f_{com}$ using learnable transformations to obtain $f_{gen}^{img}$ and $f_{gen}^{text}$ as,
\begin{equation}
    f_{gen}^{img} = \Omega^{c}_{img}(f_{com})~~~~f_{gen}^{text}  = \Omega^{c}_{text}(f_{com}) \\
\end{equation}
where $\Omega^{c}_{img}$ and $\Omega^{c}_{text}$ are learnable transformations and are trained end-to-end alongside the model. We discuss the particular impact of using this consistency loss in Section~\ref{sec:ablations:loss_functions:discriminator_loss}.

\textbf{Total Loss} used for training is computed as
\begin{equation}
    \mathcal{L}_{total} = \lambda_{1}\mathcal{L}_{triplet} + \lambda_{2}\mathcal{L}_{disc} + \lambda_{3}\mathcal{L}_{cons}
\end{equation}
$\alpha_{t}$, $\alpha_{i}$, $\beta$, $\gamma$, $\lambda_{1}$ to $\lambda_{3}$ are learnable scalar hyperparameters.

\section{Experiments}
\label{sec:experiments}
In this section, we formalize the datasets, baselines, implementation and evaluation details for our experiments. We use the same experimental and evaluation settings as used by the previous techniques to ensure consistency.

\subsection{Datasets}
We conduct experiments on multiple benchmark datasets that are selected to maximize diversity in length of the natural language descriptions. Figure~\ref{fig:dataset-statistics} (in Appendix) shows the average number of words in the support text vary from 5 to 31 across the different datasets.
\begin{figure*}[ht!]%
    \centering
    \includegraphics[page=1,width=0.75\linewidth]{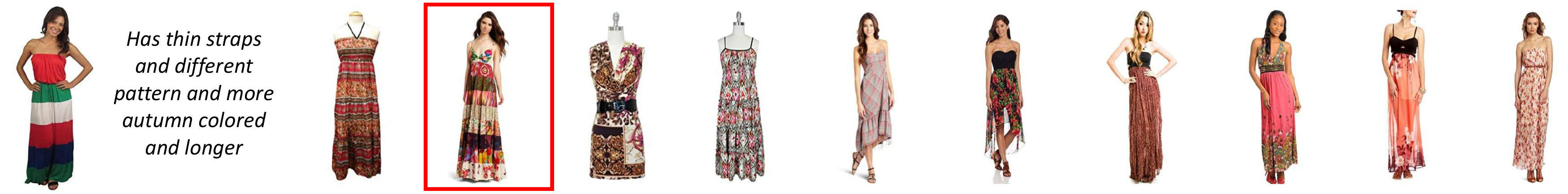}\vspace{-0.08cm}
    \includegraphics[page=5,width=0.75\linewidth]{figures/paper-additional-results.pdf}\vspace{-0.09cm}
    \includegraphics[page=9,width=0.75\linewidth]{figures/paper-additional-results.pdf}
    \caption{Qualitative results (one for each dataset) from our approach \OURNAME. Images in the first column are the reference images followed with the query text. Retrieved results are ranked from left-to-right. \textcolor{red}{Red} boxes highlight the target image.}
    \label{fig:results:qualitative-ours}
\end{figure*}
\textbf{Shoes}~\cite{showsdataset} contains 14,658 images of footwear tagged with relative captions for dialog-based interactive retrieval. The dataset is split into 10,000 training and 4,658 test images with \textit{short} support text descriptions that have an average length of 5.32 words.

\textbf{FashionIQ}~\cite{fashionIQdataset} contains 77,684 images of fashion products over 3 categories: Dress, Toptee, and Shirt, with 46,609 images in the training and 31,075 images in the validation set. The dataset is characterized by \textit{medium} support text descriptions with an average length of 10.69 words per sample. Since the ground-truth is not publicly available, so we follow VAL and report performance on the validation set.

\textbf{Birds-to-Words (B2W)}~\cite{birds2words} contains 15,931 images (12,770 training and 3,151 testing) tagged with descriptions of fine-grained differences between pairwise bird images. The natural language queries here are \textit{long} with an average length of 31.38 words.

\subsection{Experimental Setup}
\label{sec:experiments:baselines}
\textbf{Baselines}: We compare \OURNAME with a wide range of baselines including early works and recent State of the Art models on this task. \textbf{Image Only} uses only image representation as composed embedding. \textbf{Text Only} uses only text representation as composed embedding. \textbf{Concat Only} concatenates (denoted by $\mdoubleplus$) and linear transforms the image and text representations (following details from ~\cite{TIRG}) to obtain the composed embedding. \textbf{Relationship~\cite{Relationship}} takes the feature maps from final CNN layer alongwith the text feature from RNN and performs concatenation followed by MLP to learn cross-modal relationships. \textbf{FiLM}~\cite{film} is a Feature-wise Linear Modulation wherein the text information added to the CNN output to modulate each feature map by affine transformation. \textbf{TIRG}~\cite{TIRG} concatenates visual and textual representations followed by learning a gating and a residual connection to obtain a composed embedding. \textbf{VAL}~\cite{VALpaper} composes the textual representation with the visual representations at multiple CNN layers using a composite transformer (more details are mentioned in Section~\ref{sec:related_work}).

VAL is the most recent state-of-the-art technique and the strongest baseline for our experimental study. For both VAL and TIRG, we refer to the author-provided code implementations with the recommended hyper-parameter settings.

\textbf{Comparison with Workshop:} For comparison with the FashionIQ 2020/2019 Workshop (FashionIQ-W), we compare with the workshop winners who follow the standard experimental settings i.e. who do not perform pre-training of feature network on attribute prediction tasks or any other external data/tasks, and do not use an ensemble of models. We compare \OURNAME with RTIC~\cite{shin2020fashion} and CurlingNet~\cite{yu2020curlingnet}. However, RTIC uses the ResNet101 backbone while the widely used encoder for the task is ResNet50.
    
\textbf{Implementation \& Evaluation Details}: We use Resnet-50~\cite{resnet} pre-trained on ImageNet~\cite{imagenet} as the backbone for image encoder and set $L=2$ for our experiments. 
For our experiments with pretrained text embeddings, we use BERT~\cite{BERT} pretrained on QA task.
Performance is evaluated using the Recall@K (R@K) $\{K=1,10,50\}$ metric which computes the percentage of evaluation queries where the target image is found within the top-K retrieved images.
We use a batch-size of $32$ and Adam~\cite{kingma2014adam} (initial learning-rate of $1e^{-3}$) optimizer for image \& text encoders and the SGD optimizer (initial learning-rate of $2e^{-4}$) for the discriminator.
The learning rate was divided by 2 for the Adam optimizer and divided by 10 for the SGD optimizer when the loss plateaued on the validation set until it reached $1e^{-6}$.
For the image encoder, we allow for the gradual fine-tuning by unfreezing it's weights only after first few epochs (10 in our case) of training.
We use a temperature of  $\mathcal{T}={1,8}$. We choose the values $\lambda_{1}$ = 1, $\lambda_{2}$ = 0.6, $\lambda_{3}$ = 0.1 as the hyper-parameters of our loss functions.
We take $\alpha_{t}$ = 1 and $\alpha_{i}$ = 0.1 in the consistency loss.

For the discriminator $\mathcal{D}$, we use a simple neural network with three fully-connected layers that reduce the feature vectors embedding size to a scalar value that is then passed through the loss function as described in Section~\ref{sec:approach:losses}. The architecture for the Discriminator is provided in Appendix.

\subsection{Results}
We present quantitative and qualitative comparison of \OURNAME with our baselines on each of the three datasets. Due to limited space, additional results are included in the appendix which are cited where pertinent.

\textbf{Quantitative Results}: The quantitative results for all three datasets FashionIQ, Birds-2-Words and the Shoes dataset are summarized in Tables~\ref{tab:results:fashionIQ},~\ref{tab:results:birds2words}, and~\ref{tab:results:shoes} respectively. We also highlight the best number in \textbf{bold} in all the tables for convenience. We report the performance of \OURNAME using both pre-trained BERT and random embeddings. Overall, we can see that \OURNAME outperforms the strongest baseline on all three datasets by 3-4\% on average on the R@10 and R@50 metrics. Moreover, our model also outperforms the baselines on the challenging Birds to Words dataset which has much longer and more complex sentences. \footnote{Results on B2W dataset for VAL were not available and experimenting using their code was prohibitive even with 16-GB GPUs}

\textbf{Qualitative Results}: To corroborate our quantitative observations, we also present a qualitative analysis for \OURNAME and present the results for the same in Figure~\ref{fig:results:qualitative-ours}. We observe that the \OURNAME is able to concurrently incorporate multiple semantic transformations in visual representations from text descriptions when retrieving images.
We observe that \OURNAME is able to -- \textbf{(A)} retrieve new images while changing certain attributes conditioned on text feedback eg. color, material (from row 1, \OURNAME captures the ``autumn-colored'' while preserving the ``longer'' property) \textbf{(B)} ingest multiple visual attributes and properties in the natural language text feedback (from row 2, ``solid'', ``black'' and ``orange-and-beige trim'' all focus on different semantics of the image. \OURNAME captures all of the semantics in the retrieved image) \textbf{(C)} can jointly comprehend global appearance and local details for image search (from row 2, \OURNAME captures the overall ``black'' look across the retrieved results and attempts to find the appropriate local variations in the design) \textbf{(D)} aggregate multiple fine-grained semantic concepts within query sentence for image search (from row 3, \OURNAME captures the fine-grained changes like ``black breast'', ``green flashes'' and ``longer beak'' in a single query and aggregates these concepts effectively).
Due to limited space, we have included comparative qualitative analysis with VAL and additional qualitative results in Appendix~\ref{appendix:qualitative-res}.

\begin{table}[]
    \centering
    \scriptsize
    \begin{tabular}{|c|c|c|}
        \hline
        \textbf{Composition}         & \textbf{R@10}           & \textbf{R@50}           \\ \hline
        Concatenation       & 22.98          & 47.73          \\ \hline
        Residual Gating     & 24.00          & 46.72          \\ \hline
        Hadamard            & 22.74          & 46.35          \\ \hline
        Residual Offsetting & \textbf{27.83} & \textbf{54.02} \\ \hline
    \end{tabular}
    \caption{\label{tab:hfm}%
    Results on ablations for effect of composition}
\end{table}
\begin{table}[]
    \centering
    \scriptsize
    \begin{tabular}{|c|c|}
        \hline
        \textbf{Method}    & \textbf{\# of Parameters} \\ \hline
        TIRG      & 16.96M           \\ \hline
        ComposeAE & 19.07M           \\ \hline
        VAL       & 61.76M           \\ \hline
        Ours      & 21.74M           \\ \hline
    \end{tabular}
    \caption{%
    \label{table:num_parameters}%
    Number of params for different methods}
\end{table}
\begin{table}[]
    \centering
    \small
    \begin{tabular}{|c|c|c|}
        \hline
        \multirow{2}{*}{Loss Functions} & \multicolumn{2}{c|}{Average} \\ \cline{2-3}
            & R@10 & R@50 \\ \hline
        $\mathcal{L}_{triplet}$ & 23.47 & 48.48 \\
        $\mathcal{L}_{triplet}$ + $\mathcal{L}_{cons}$ & 24.82 & 50.25 \\
        $\mathcal{L}_{triplet}$ + $\mathcal{L}_{disc}$ & 22.84 & 48.87\\ \hline
        $\mathcal{L}_{triplet}$ + $\mathcal{L}_{disc}$ + $\mathcal{L}_{cons}$ & \textbf{27.83} & \textbf{54.02} \\ \hline
    \end{tabular}
    \caption{Results from our ablation study showcasing the impact of individual losses on FashionIQ dataset.%
    \label{tab:ablations:losses:detailed_numbers}}
\end{table}

\begin{table}[]
    \centering
    \scriptsize
    \begin{tabular}{|c|c|c|c|}
        \hline
        \textbf{Level-1} & \textbf{Level-2} & \textbf{R@10} & \textbf{R@50} \\ \hline
        \checkmark       &                  & 8.95          & 24.13         \\ \hline
                         & \checkmark       & 14.24         & 31.05         \\ \hline
        \checkmark       & \checkmark       & 27.83         & 54.02         \\ \hline
    \end{tabular}
    \caption{Ablations for aggregating both levels}
    \label{tab:levels}
\end{table}

\subsection{Ablation Studies}
\label{sec:ablations}
In this section, we conduct ablation studies to investigate the impact of different design choices in \OURNAME. For all our ablations, we restrict our scope to the FashionIQ dataset for ease of exposition and analysis.

\begin{figure}[htb!]
    \centering
    \includegraphics[page=1,width=1\linewidth]{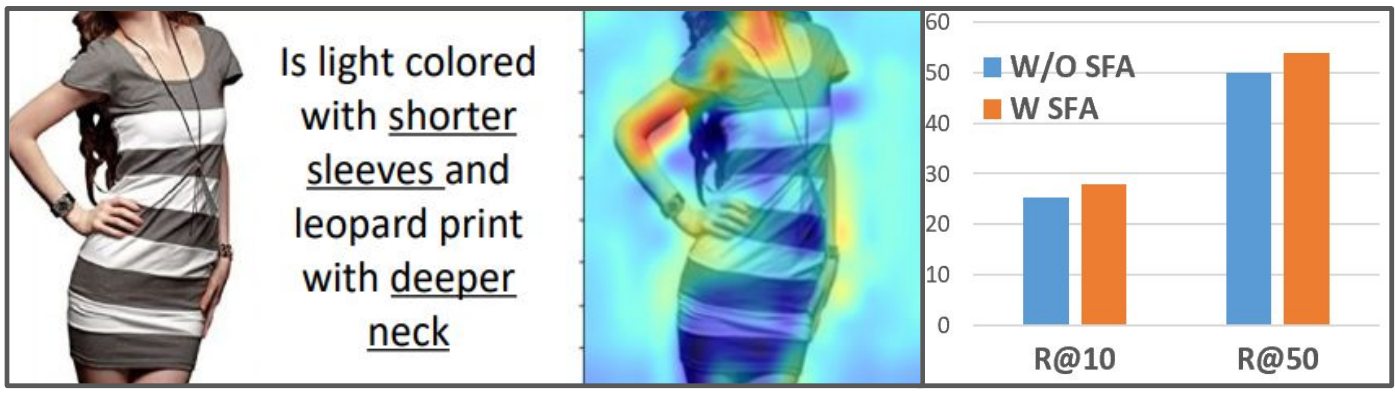}
    \caption{On the left, Attention maps for a pair of input image and text with specific keywords \underline{\textit{underlined}} corresponding to the attention heat-maps on the image. On the right, the effect of SFA is shown on the Fashion-IQ dataset}
    \label{fig:attention_maps}
\end{figure}

\textbf{Importance of SFA and Attention Maps}
\label{sec:attention_maps}
We provide the attention map from the last level ($\mathbf{\mathcal{A}}^{\ell}_{\mathbf{cross}}$) from \OURNAME in Figure~\ref{fig:attention_maps}. From the figure, the network focuses on the region of the sleeve in the image since the text has ``shorter sleeves'', and the neck region as the text said ``deeper neck''. We also provide additional attention maps in Figure~\ref{fig:attention_maps:additional} in Appendix~\ref{appendix:attention_maps}. Besides, we also show the importance of the SFA module on the right side of Figure ~\ref{fig:attention_maps} and it can be seen that adding the SFA module improves the R@10 and R@50 metrics by around 3\%. 
Since, for this problem at hand and our method, it wouldn't be logical to run analysis by removing SFM (`how to change'), which would make it equivalent to Image Only baseline as discussed in Section~\ref{sec:experiments:baselines}, hence, we omit the same in our study. \\
\textbf{Effect of Residual Offsetting in SFM:}
\label{sec:ablations:composed_representations:residual_composition} 
Here, we study our idea to utilize text to only “modify” the image feature based on the text feature, rather than create an entirely.
Correspondingly, we validate this design choice by contrasting with the following operators: \textit{Concatenation}, \textit{Hadamard Product}, \textit{Residual Gating} (used in TIRG) and \textit{Residual Offsetting} (defined in Eq.~\ref{eq:residual-composition}). The results are summarized in Table~\ref{tab:hfm} which highlights that our operator significantly outperforms the alternate choices. \\
\textbf{Importance of Aggregating both Levels:}
Here, we study the effect of aggregating both the levels in contrast to using one of them. Table~\ref{tab:levels} shows how taking coverage of concepts over both the levels results in better performance. \\
\textbf{Effect of Discriminator and Consistency Loss:}
\label{sec:ablations:loss_functions:discriminator_loss}
\textbf{Discriminator loss} is defined to provide a weaker supervision to further knit the two distributions ($f_{com}$ and $f_{tgt}$) together while \textbf{Consistency Loss} is designed to regularize the learned composite multi-modal representations (see Section~\ref{sec:approach:losses}) Table \ref{tab:ablations:losses:detailed_numbers} shows the efficacy of the two loss functions. \\
\textbf{Comparison of Number of parameters}: Proposing a simple yet efficient approach, we compare the number of parameters  against existing SOTA (VAL~\cite{VALpaper}, TIRG~\cite{TIRG} and ComposeAE~\cite{composeAE}) in Table \ref{table:num_parameters}. Table~\ref{tab:results:fashionIQ} shows how our model outperforms VAL significantly, while using just one-third number of parameters. Moreover, by adding just 14\% parameters to ComposeAE, our model achieves a gain of 10\% on R@10 metric.


\section{Conclusion and Future Work}
\label{sec:conclusion}
In this work, we focus on the task of text conditioned image retrieval and introduce \OURNAME, which resolves the given task into 2 major steps, SFA (where to see) and SFM (how to change) which systematically streamlines the generation of text aware image features. We conduct extensive experiments on diverse benchmark datasets and consistently achieve state-of-the-art performance. 

There are some cases when all the predictions are qualitatively coherent but this is not captured by metric since the specific target image is not a part of the retrieved set.
Thus, exploring adaptive evaluation metrics is an interesting direction for future work.



{\small
\bibliographystyle{ieee_fullname}
\bibliography{egbib}
}

\clearpage

\appendix

\section*{\textit{Appendix for}\\ \Title}
\label{appendix}
Our supplementary material is organized as follows: In Appendix~\ref{appendix:implementation_details}, we provide implementation details for training the \OURNAME models. 
We provide qualitative results on this task in Appendix~\ref{appendix:qualitative-res} with a detailed analysis of the results obtained.

\section{Implementation Details}
\label{appendix:implementation_details}
We conduct our experiments on a machine with $4 \times TESLA~V-100$ GPU's, $16 \times CPU$ cores with 2 threads per-core and 16 GB RAM installed with Ubuntu 16.04 LTS operating system. We used the Pytorch==1.14.0 framework for our experiments.
The architecture of the Discriminator $\mathcal{D}$ used in \OURNAME is provided below
\begin{center}
\begin{tabular}{c}
\begin{lstlisting}[linewidth=\columnwidth,breaklines=true,language=python]
nn.Sequential(nn.Linear(512,512),
nn.LeakyReLU(0.2,inplace=True),
nn.Linear(512,256),
nn.LeakyReLU(0.2,inplace=True),
nn.Linear(256,1),
nn.Sigmoid())
\end{lstlisting}
\end{tabular}
\end{center}


\section{Attention maps}
\label{appendix:attention_maps}
\begin{figure}[hbt!]
    \centering
    \includegraphics[page=2,width=0.6\linewidth]{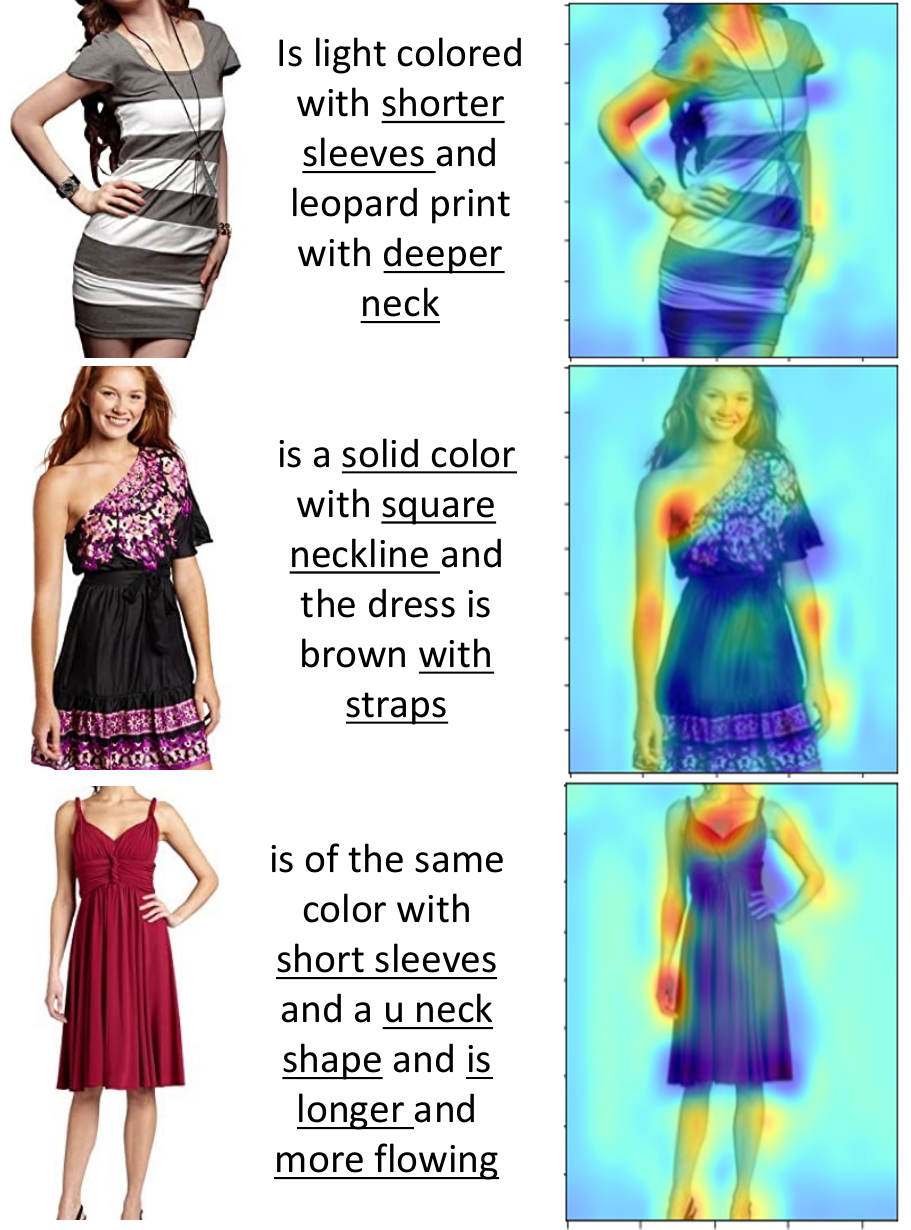}
    \caption{Attention maps for the different pairs of input images and text with specific keywords \underline{\textit{underlined}} based on the attention heat-maps.}
    \label{fig:attention_maps:additional}
\end{figure}

We provide additional attention maps for different images in Figure~\ref{fig:attention_maps:additional}.
\begin{figure}[bth!]
    \centering
    \includegraphics[page=1,width=0.75\linewidth]{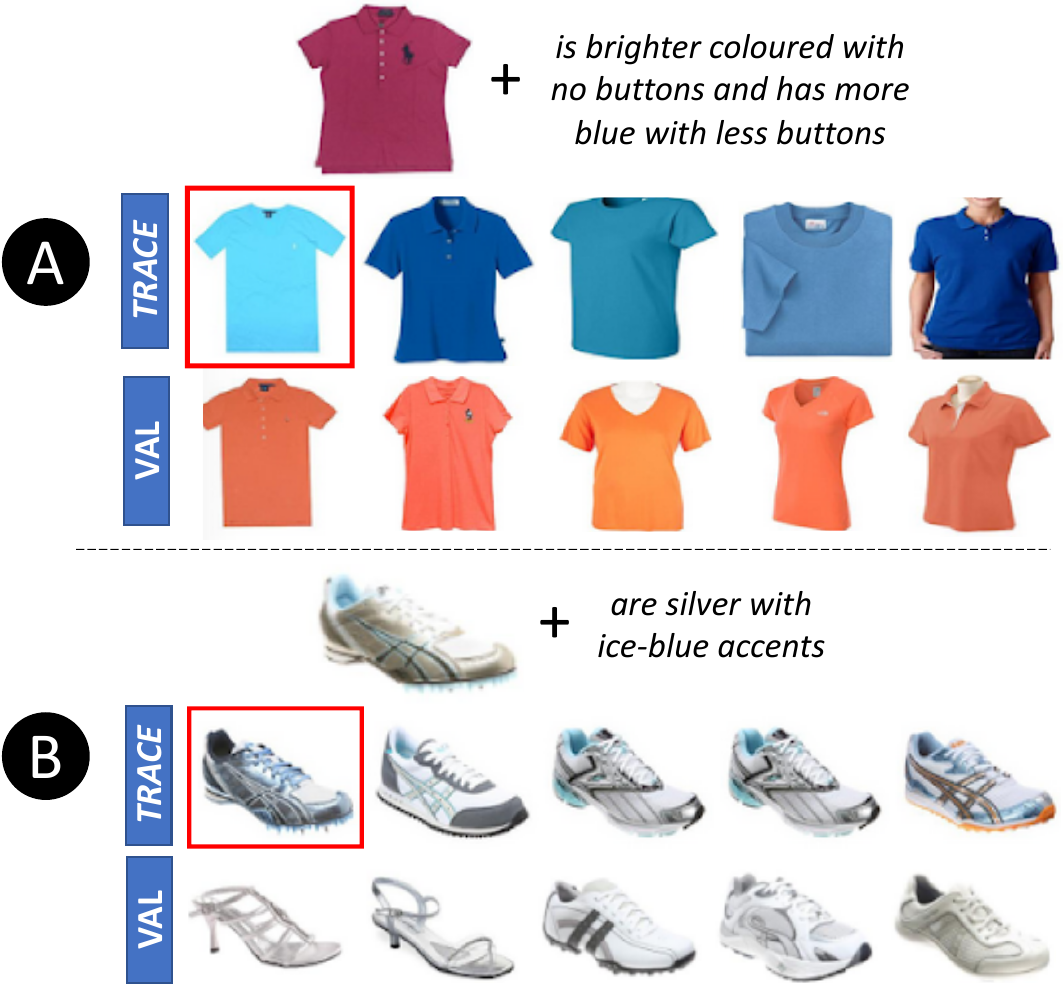}
    \caption{Qualitative results from \OURNAME compared with VAL. First row shows the reference image and the query text. \textcolor{red}{Red} boxes denote the correct target image}
    \label{fig:results:qualitative_comp}
\end{figure}


\section{Additional Qualitative Results}
\label{appendix:qualitative-res}
\begin{figure*}[thb!]
    \centering
    \includegraphics[page=2,width=0.85\linewidth]{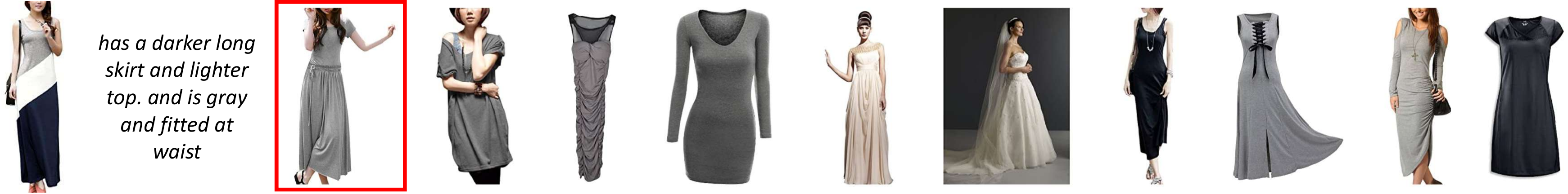}\\
    \includegraphics[page=3,width=0.85\linewidth]{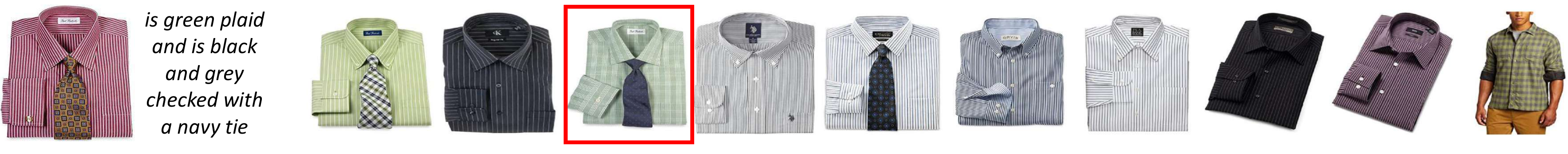}\\
    \includegraphics[page=4,width=0.85\linewidth]{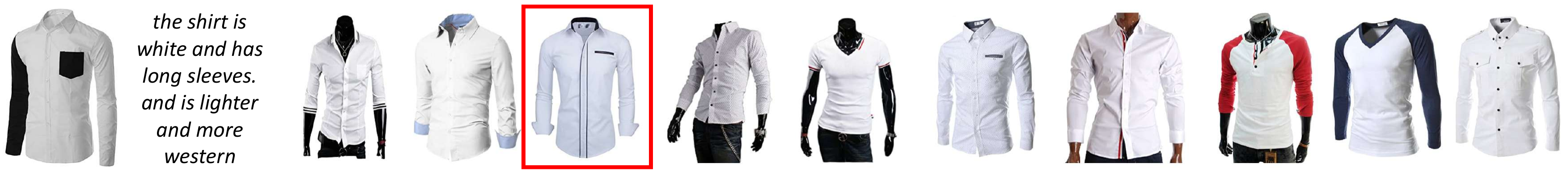}\\
    \includegraphics[page=6,width=0.85\linewidth]{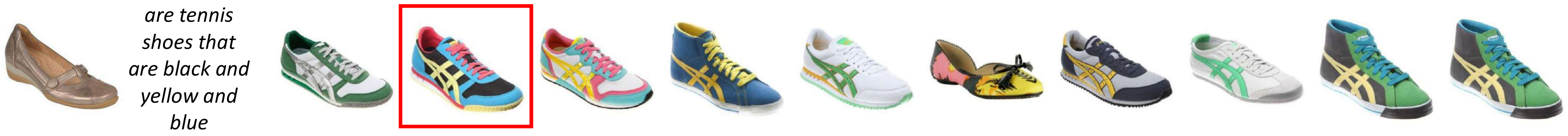}\\
    \includegraphics[page=7,width=0.85\linewidth]{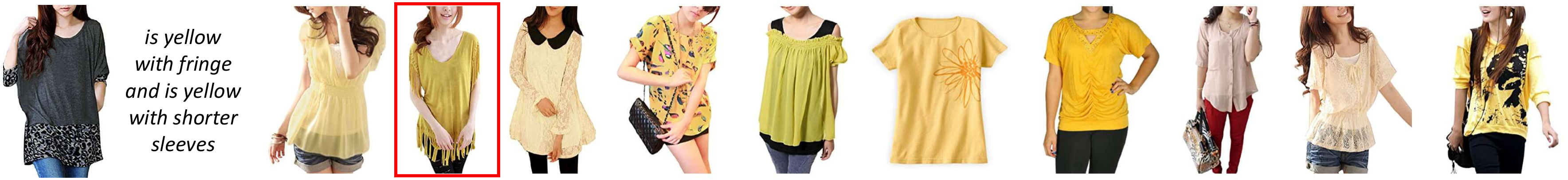}\\
    \includegraphics[page=8,width=0.85\linewidth]{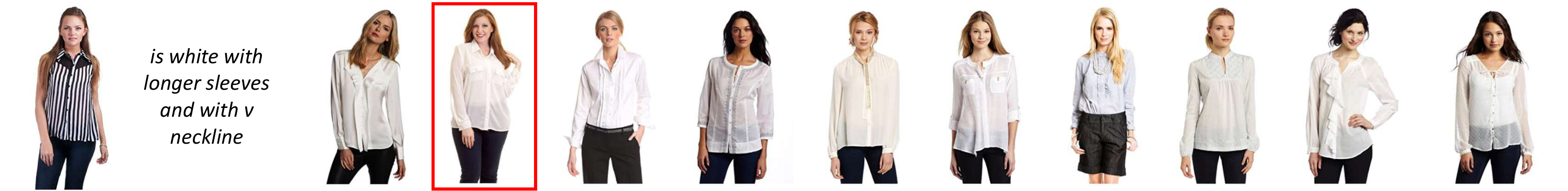}\\
    \includegraphics[page=10,width=0.85\linewidth]{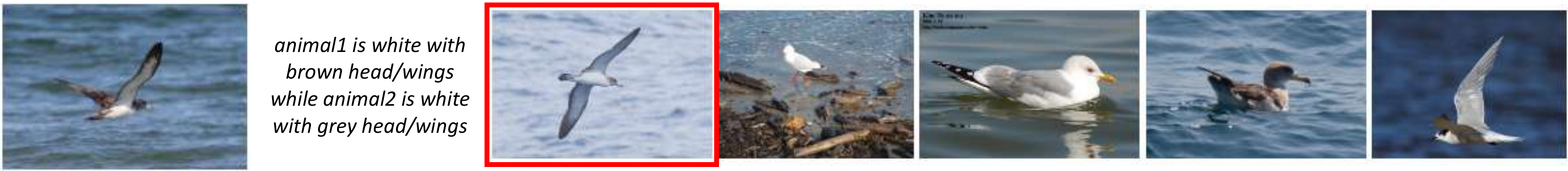}
    \caption{\label{fig:results:additional_results}%
    Additional Qualitative results for our proposed method \OURNAME on FashionIQ, Shoes, and Birds-to-Words datasets. The first column denotes the source image followed by the support text description. The retrieved results then follow with ranking in decreasing relevance from left-to-right.}
\end{figure*}
\begin{figure*}[th!]
    \centering
    \includegraphics[page=1,width=0.7\linewidth]{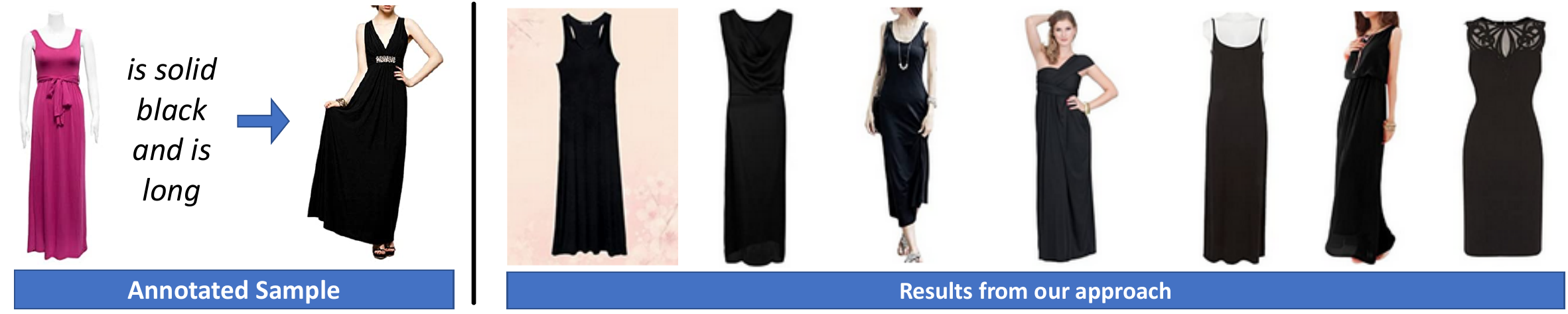}
    \label{fig:ablations:label_not_found}
    \caption{On the left, we provide the query image and the corresponding textual query along with the ground truth from the FashionIQ dataset. On the right, we provide the retrieved results ranked by decreasing relevance results from our approach \OURNAME. Though the retrieved results don't contain the actual annotated sample, the results are extremely relevant to the query}
\end{figure*}
Figure~\ref{fig:results:additional_results} presents additional qualitative results for \OURNAME on each of the three datasets.

In Figure~\ref{fig:results:qualitative_comp}, we present comparative visual results for \OURNAME with VAL, the strongest baseline, on the FashionIQ and Shoes datasets. In particular, we notice that \OURNAME is able to i) retrieve images while changing cross-granular attributes conditioned on text feedback (from Figure~\ref{fig:results:qualitative_comp} (A), \OURNAME retrieves shirt which is ``blue colored'' with ``less buttons'' while VAL incorrectly retrieves ``orange'' colored clothes) ii) concurrently focus on global appearance and multiple local fine-grained designs (from Figure~\ref{fig:results:qualitative_comp} (B), \OURNAME retrieves ``\underline{shoes}'' that have ``silver color'' and ``ice blue accents'' while VAL incorrectly retrieves sandals).

\end{document}